\pdfoutput=1

\documentclass[11pt]{article}

\usepackage[]{acl}

\usepackage{times}
\usepackage{latexsym}

\usepackage[T1]{fontenc}

\usepackage[utf8]{inputenc}

\usepackage{microtype}

\usepackage{inconsolata}

\usepackage{graphicx}
\usepackage{amsmath} 
\usepackage{booktabs} 
\usepackage{array}
\usepackage{makecell}
\usepackage{multirow}
\usepackage{multicol}
\usepackage{bm}

\usepackage{_macros}
\usepackage{float}

\usepackage[most]{tcolorbox}
\usepackage{xcolor}

\definecolor{PineGreen}{rgb}{0.0,0.47,0.44}
\definecolor{BrickRed}{rgb}{0.8, 0.25, 0.33}
\definecolor{ForestGreen}{rgb}{0.0, 0.57, 0.13}
\definecolor{Red}{rgb}{1.0, 0.0, 0.0}

\newtcolorbox{promptbox}{
  colback=gray!8,
  colframe=gray!60,
  boxrule=0.48pt,
  arc=1mm,
  left=2mm, right=2mm, top=1mm, bottom=1mm,
  breakable,
  enhanced,
  halign=flush left,
  fontupper=\ttfamily\small,
}

\title{\methodname: Leveraging Unlabeled Text via \\ Self-Relevance-Guided Masking for Decoder-Only Classification}

\author{\\
 \textbf{Pietro Ferrazzi\textsuperscript{1,2}},
 \textbf{Matteo Merler\textsuperscript{1}}, \\
 \textbf{Giovanni Bonetta \textsuperscript{1}},
 \textbf{Alberto Lavelli \textsuperscript{1}},
 \textbf{Bernardo Magnini\textsuperscript{1}}
\\
 \textsuperscript{1}Fondazione Bruno Kessler, Trento, Italy\\
 \textsuperscript{2}University of Padova, Italy \\
  \small{
    \textbf{Correspondence:} \href{mailto:pferrazzi [at] fbk [dot] eu}{pferrazzi [at] fbk [dot] eu}
  }
}

\begin{document}
\maketitle
\begin{abstract}
Classification tasks require annotated data, which can often be expensive, time-consuming, or even unfeasible to collect.
This is the case of the medical domain, where large datasets often have few annotated examples. 
To address this, we propose DecSelfMask (Decoder Self-learning by Masking), an approach to enhance decoder-only performance on classification tasks. 
We build on common self-learning approaches by leveraging a model to create training examples from unlabeled data, and propose a novel relevance-guided masking strategy.
We use relevance attribution methods to determine what portions of unannotated texts are relevant for a task. 
We then create self-supervised training examples by masking out those portions, training the model to reconstruct them via next-token-prediction. We hypothesize that those examples convey knowledge about the structure and semantics of unannotated data that can be useful for downstream performance. 
We test our approach on 136 tasks from a collection of 1.9M clinical notes from an Italian hospital. 
We quantify DecSelfMask's impact on 
downstream tasks on 5 models of different scales and families, including a probing analysis. 
Experiments show consistent gains, outperforming the base models (+9.1 points in Macro F1), continual pretraining (+6.3), synthetic label generation (+12.5), as well as common baselines. 
The results show that relevance attribution can serve as a powerful source of supervision for decoder learning from unlabeled data, beyond existing methods.
\end{abstract}

\section{Introduction}

\begin{figure*}[t]
  \includegraphics[width=\textwidth]{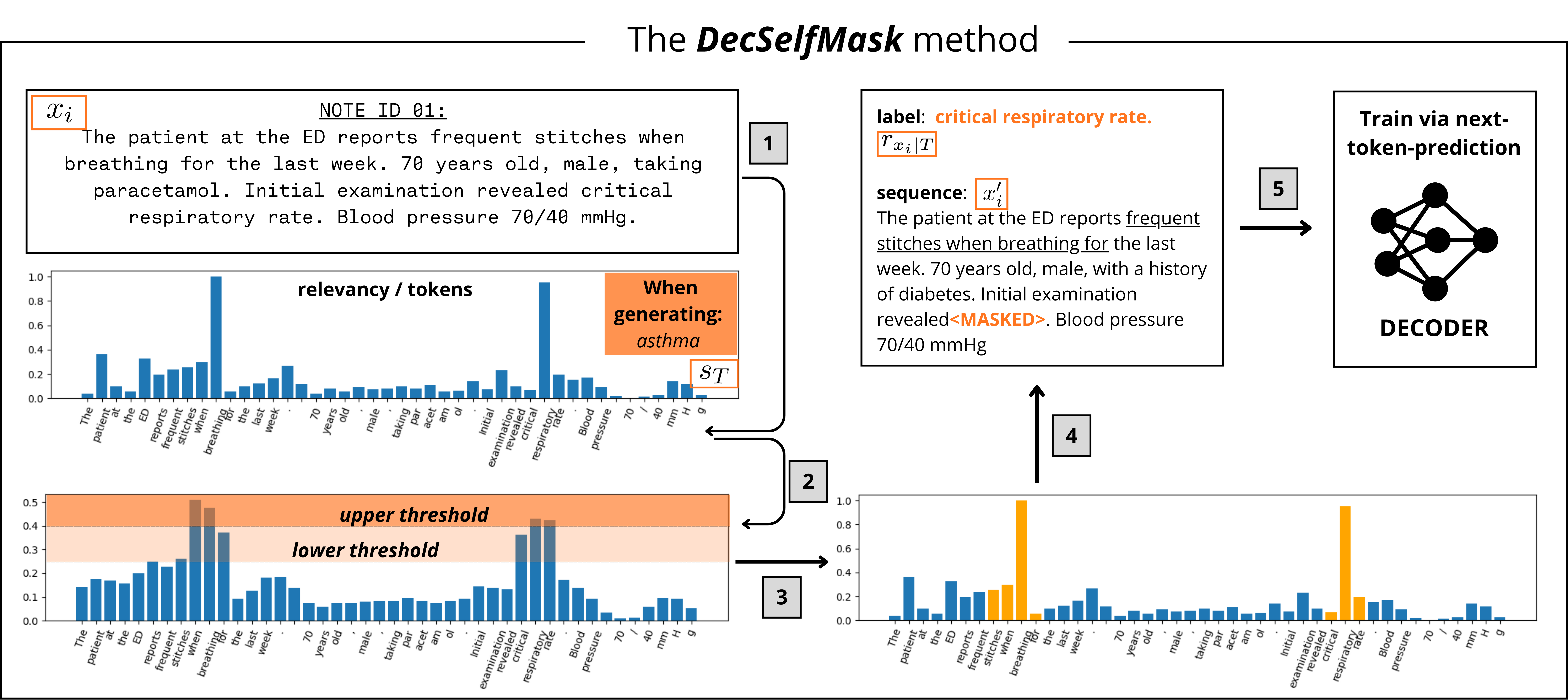}
  \caption{
      \textbf{\methodname method overview}, exemplified by the task of determining if a patient has asthma. 
  \textbf{1) Relevancy calculation}: the raw text $x_i$ is pre-pended to the target text portion $s_T$ (e.g., \textit{asthma}) and given as input to the model. A relevance score is computed for each input token with respect to $s_T$.
  \textbf{2) Gaussian smoothing}: the relevance scores are smoothed and two thresholds are identified.
  \textbf{3) Groups generation}: portions of relevant text $r_{x_i|T}$ are identified by all consecutive tokens with relevance higher than a lower threshold, where at least one token has relevance higher than the lower threshold.
  \textbf{4) Masking}: training sequences $x_i'$ are generated by masking out relevant portions of text $x_i$.
  \textbf{5) Training}: decoders are trained via next-token-prediction on the generated examples.
  }
  \label{fig:training_data_creation}
\end{figure*}

Training automatic systems to perform Natural Language Processing tasks requires large amounts of high-quality annotated data~\citep{vanEngelen2020, YAKIMOVICH2021100383}. 
Annotations collection  can be performed by non-experts~\citep{snow-etal-2008-cheap}, but there are cases where high degrees of expertise are required~\citep{yang-etal-2019-predicting}. Such processes can be expensive and time-consuming, and in domains like medicine they can represent a major issue~\citep{YAKIMOVICH2021100383}. 

Because of these challenges, medical applications often involve an abundance of raw data, and limited annotated examples.
Common approaches to tackle these scenarios rely on \textit{self-learning} ~\citep{wang-etal-2023-self-instruct}, where manual annotations are first leveraged to generate synthetic ones, which are then combined with the former to train downstream models ~\citep{shi-etal-2023-rethinking,luo-etal-2025-semi}. 
Other approaches use raw data to perform \textit{continual pretraining}, aiming to adapt models to the target domain \citep{Singhal2023,Wu2024}, building on the implicit relation existing between the raw text  and the  downstream task.

While \textit{self-learning} tries to project such knowledge into unannotated text and expand the task-oriented training set, \textit{continual-pretraining} attempts to gain domain-oriented knowledge from the data itself, without explicitly considering the target tasks.
An interesting direction remains underexplored at their intersection, combining the acquisition of knowledge from raw data and the guidance that can be given by annotated examples. 
For instance, a target task on \textit{"asthma"} classification can provide guidance to automatically identify portions of a raw text such as \textit{"frequent stitches when breathing"} and \textit{"critical respiratory rate"}, to learn the implicit relation existing between them, and potentially increase downstream performance on the target task itself (Figure~\ref{fig:training_data_creation}).

Following this hypothesis, we propose \textbf{Decoder Self-learning by Masking} (\methodname), a method to create training examples from the knowledge embedded in unannotated data. At its core, it is a self-learning-based technique to automatically generate masked examples to train decoder-only models, specifically targeting relevant portions of the input.
Our contribution is a theoretical framework and the experimental evidence that relevance attribution can itself serve as a powerful source of supervision for decoder learning from unlabeled data. It can be articulated as follows:
\begin{itemize}
    \item We propose \methodname, a method to combine training-based domain adaptation (continual pretraining)  with self-supervised learning, aiming to increase performance on classification tasks.
    \item We evaluate \methodname on medical $136$ tasks, finding an increase from standard approaches of $+19.9$ Macro F1 points on average among models scales and families.
    \item We compare \methodname to several baselines and state-of-the-art approaches, with gains of $+6.3$ over continual pretraining and $+12.5$ over self-learning.
\end{itemize}
We release the codebase to utilize our framework on custom data and replicate the experiments\footnote{\url{https://github.com/ferrazzipietro/DecSelfMask}}.

\section{Related Work}
\label{sec:related_work}


\paragraph{Semi-supervised Learning and Self-training.}
Automatic systems can be \textit{supervised} or \textit{unsupervised}~\cite{bishop2006pattern}.
When labeled data is present but scarce, mixed \textit{semi-supervised} approaches have shown to be effective~\citep{zhu2008semisupervised}. 
With the rise of LLMs, these methods have shifted toward \textit{self-learning} \citep{Singhal2023}, where models are trained on the labeled examples aiming to generate synthetic-labels from unannotated data, to be then utilized for further training~\citep{gera-etal-2022-zero,shi-etal-2023-rethinking, wang-etal-2023-self-instruct, luo-etal-2025-semi}. 
These methods consist in bootstrapping model's own knowledge about the downstream task.
By contrast, we use models' own estimates to automatically identify task-relevant information in raw text and transform it into self-supervised training signal, without explicit reference to the task itself.

\paragraph{Medical Domain Adaptation via Continual Pretraining.}
With the advent of LLMs, there has been a growing interest in models adaptation to the medical domain. The main technique is continual pretraining (CPT) \citep{ou-etal-2025-llms}, where models are exposed to large corpora of raw medical data via next-token-prediction, as this requires no expert labels. Examples are BioGPT~\citep{Luo_2022},  MedPALM-1~\citep{Singhal2023}, PMC-LLaMA~\citep{Wu2024}, MedPalm-2~\citep{Singhal2025}, MedPhi~\citep{corbeil-etal-2025-modular}, MedGemma~\citep{sellergren2026medgemmatechnicalreport}. These methods have shown to be effective by aligning models' knowledge to the target domain, without directly modeling downstream tasks. Notably, others have shown CPT having low-to-no impact on downstream medical tasks~\citep{ferrazzi2026smallllmsmedicalnlp, belmadani-etal-2026-trade} suggesting that alternative approaches should be explored. We follow this line by introducing task-informed signal to CPT.  


\paragraph{Decoder-only Masking Strategies.}
\citet{dukic-snajder-2024-looking, HUANG2025112907} propose to act on decoders' causal mask to improve performance on sequence labeling and question answering tasks.
More similar to us,~\citet{khosla-etal-2025-magnet} present a method to enable decoders to generate sentence-level representations 
to perform text in-filling using bi-directional context, yet not performing any self-learning nor task-oriented optimization.
By contrast, we do not modify the models' structure in any way.

\paragraph{Relevance Attribution in Transformers.}
Several methods have been proposed to determine the relevance of the input for a given output generated by a transformer-based model
\citep{ferrando2024primerinnerworkingstransformerbased}. While perturbation approaches~\citep{JMLR:v22:20-1316} and contrastive attribution~\citep{yin-neubig-2022-interpreting} require extensive comparisons and alternative search, gradient-based methods can be applied independently from the context. Hence, we consider AttnLRP~\citep{pmlr-v235-achtibat24a} for input relevance attribution in the context of our method.

\section{Problem Formulation and Hypothesis}
\label{sec:problem_formulation}

In this work, we propose a method to leverage the 
structure of unlabeled,
domain-specific corpora to solve downstream classification tasks.
Our approach relies on the hypothesis that if relevant information can be automatically identified, it can be exploited to construct self-supervised learning objectives, aligned with the downstream task.

\paragraph{Learning from Unlabeled Data.}
Let us define $X = \{x_i\}_{i=1}^{N}$ denoting a collection of sequences sampled from a domain-specific corpus, and let $C=\{c_i \in \{c_1, ..., c_K\}\}_{i=1}^{N}$ denote the relative class labels for the classification task $T$, which are unknown.
Following~\citet{vanEngelen2020}, we investigate whether the marginal distribution over the input space $\operatorname{P}(x)$ contains useful information for learning the posterior distribution $\operatorname{P}(c \mid x)$.  In our setting, a decoder-only language model $\operatorname{LLM}(c \mid x)$ is used to estimate this conditional distribution.
Our central hypothesis, similar to the one behind continual pretraining, is that the conditional distribution $\operatorname{P}(c \mid x)$ is not independent from the structure of the input distribution $\operatorname{P}(x)$. Consequently, we hypothesize that learning regularities, dependencies, and semantic structures within the unlabeled examples sampled from $\operatorname{P}(x)$ can improve performance on the downstream task $T \sim \operatorname{P}(c \mid x)$.
However, the relationship between $\operatorname{P}(x)$ and $\operatorname{P}(c \mid x)$ is generally implicit and task-dependent~\citep{NEURIPS2018_c1fea270}. In what follows, we describe our approach to verify this hypothesis.

\paragraph{Linking $\mathbf{\operatorname{P}(c|x)}$ and $\mathbf{\operatorname{P}(x)}$ through  $\operatorname{LLM}$.}
Classification tasks rely on the fact that given an input sequence $x_i$ and a task $T$, there are portions of $x_i$ that are relevant to predict the task-related label $c_i$. Let
\begin{equation}
r_{x_i|T}(\text{start},\text{end}) = \{x_{i,j}\}_{j=\text{start}}^{\text{end}} \subset x_i
\end{equation}
denote a subset of tokens in $x_i$ identified as \textit{relevant}\footnote{For simplicity, in the following we refer to $r_{x_i|T}(\text{start},\text{end})$ as $r_{x_i|T}$} for the downstream task $T$. 
Instead of directly learning $\operatorname{LLM}(c_i \mid x_i)$, 
we define an auxiliary self-supervised objective:
\begin{equation}
\label{eq:self-sup}
\operatorname{\operatorname{LLM}}(r_{x_i|T} \mid x_i'),
\end{equation}
where the model predicts a relevant portion of the input $r_{x_i|T}$ from $x_i' = x_i \setminus r_{x_i|T}$, the input text itself where the relevant portion has been masked out (Figure~\ref{fig:training_data_creation}, top right). This formulation resembles self-learning, where the model itself identifies the labels to learn from.
Our hypothesis is then framed as follows: the $\operatorname{LLM}$ can be trained on the objective defined in Equation~\ref{eq:self-sup} (which solely depends on the input $\operatorname{P(x)}$) 
to better approximate $\operatorname{P(c|x)}$.

In other words, we hypothesize that learning to auto-regressively predict task-relevant portions $r_{x_i|T}$ from $x_i'$ encourages the model to acquire domain-specific knowledge, linguistic structures, and latent correlations that are also beneficial for the downstream task $T$.

\paragraph{The "Two Peaks" Intuition.}

We aim to train models on those inputs $x_i$ that convey some relevant knowledge $r_{x_i|T}$ about the downstream task $T$, according to the structure of Equation~\ref{eq:self-sup}. On the other hand, we rely on utilizing $r_{x_i|T}$ as a label that needs to be predicted, masking it out from the training sequence $x_i'$. Then, there is no guarantee that, after masking, $x_i'$ will convey enough task-relevant information to properly predict $r_{x_i|T}$ in the context of $T$. 
Therefore, we include in our training set only sequences with two or more disjointed relevant  portions $r_{x_i|T}(s_1,e_1)$ and $r_{x_i|T}(s_2,e_2)$ (Figure~\ref{fig:training_data_creation}, bottom right). By doing so, we leverage the self-learning framework to its full extent, by training models to predict one portion of text that the model itself identified as relevant to the task, ensuring at least another one exists to provide enough context for this to be possible. We refer to this idea as the "two peaks" framework, given that we only select sequences with at least two peaks of relevance.

\paragraph{Relevance Estimation.}
Our method relies on the ability to extract portions $r_{x|T}$ that are somehow \textit{relevant} for the downstream task $T$, to then expose models to that information. Following the concept of \textit{self-learning}, we compute such relevance by leveraging the model itself. 
We append to each input $x_i$ a short string $s_T$ that describes the task $T$ (in the example of Figure~\ref{fig:training_data_creation}, $s_T=``asthma``$).
Then, we use AttnLRP~\citep{pmlr-v235-achtibat24a} to quantify the relevance attributed by the model itself to each token in $x_i$ if it was generating $s_T$. For a given sequence of input tokens $x_i=(x_{i,1},...x_{i,n})$,  AttnLRP provides a relevance score for each token 
$\operatorname{rel}_{x_i|s_T} = (\operatorname{rel}_{x_{i,1}|s_T},\dots,\operatorname{rel}_{x_{i,n}|s_T})$.

\paragraph{Training Data Selection.}

We define a sequence $x_i$ as relevant for a task $T$ if there exist at least two disjoint subsequences where all tokens have relevance higher than a lower threshold $t_{l}$, and at least one token has relevance higher than an upper threshold $t_{u}$ (following the "two peaks" framework). Formally, each of the two disjoint subsequences $r_{x_i|T}(s,e)$ complies with the following conditions:

\begin{equation}
\small
\label{eq:relevant_span}
\begin{aligned}
\operatorname{cond}_1(s,e) &:
G(\operatorname{rel}_{x_{i,j}|s_T}) \geq t_l
\quad \forall j \in [s,e]
\\
\operatorname{cond}_2(s,e) &:
\exists j \in [s,e]
\;\text{s.t.}\;
G(\operatorname{rel}_{x_{i,j}|s_T}) \geq t_u,
\end{aligned}
\end{equation}
where $G()$ is a gaussian smoothing function on the relevance scores (Figure~\ref{fig:training_data_creation}, step 2).



Since $\operatorname{rel}_{x_{i,j}|s_T}$ is defined over tokens, $r_{x_i|T}$  spans may not align with full words and can split sub-word tokens. For smoothing purposes, we extend each span to cover the complete words intersecting the selected token range.

The objective proposed in Equation~\ref{eq:self-sup}, combined with the training data defined in this manner, acts as a form of task-oriented self-supervision: the model extracts supervision signals directly from unlabeled data by leveraging the fact that it has identified as relevant for the task at least another portion of text.

\section{Experimental Setup}
\label{sec:setup}

\begin{table*}[ht]
  \centering
  \small
  \begin{tabular}{lp{5.5cm}p{4cm}ccc}
    \toprule
     \textbf{Dataset} &    \multicolumn{1}{c}{\textbf{Description}} & \textbf{Use} & \textbf{n} & \textbf{words} & \textbf{n tasks} \\ \midrule
     SGB & Anonymized raw clinical notes from the emergency department of an Italian hospital. & Generation of the \methodname training examples. & 1.9M & 125.7M & \textemdash{} \\ \midrule
     CRF & Annotated clinical notes from the SGB dataset on 136 medical items. Each note is annotated once for each relevant item.  & Guidance on how to construct the \methodname training examples; main evaluation. & 61k & 4.7M & 136\\ \midrule
     Chronicity & Annotated clinical notes from the SGB dataset determining whether a patient has chronic conditions. & Evaluate \methodname on tasks not involved in its definition and training. & 2587 & 335k & 1  \\\midrule
     T-D & Annotated clinical notes from the SGB dataset determining whether a patient is admitted to the emergency department with loss of consciousness or dyspnea. & Evaluate \methodname on tasks not involved in its definition and training. & 1713 & 331k & 1 \\
    
\bottomrule
  \end{tabular}
  \caption{\textbf{Datasets used in our experiments}, with number of examples \textit{n}, \textit{words}, and \textit{tasks} (where applicable). \textit{SGB} is used for \methodname learning, to improve performance on classifying notes on the 136 \textit{CRF} items. \textit{Chronicity} and \textit{T-D} are not directly involved in the \methodname process, but left for further testing. }
  \label{tab:data}
\end{table*}

To verify the hypothesis defined in Section~\ref{sec:problem_formulation}, we require a large dataset of \textbf{unlabeled data} together with a smaller set of \textbf{expert-annotated data} for medical classification tasks. We leverage the SGB dataset of medical text presented by~\citet{ferrazzi-etal-2026-small}, released with a permissive license. This comprises around 1.9 million anonymized clinical notes from the Emergency Department of the San Giovanni Bosco Hospital (Turin, Italy). 

\citet{ferrazzi2026crf} further proposed the Case Report Form (CRF) filling task built on top of the SGB dataset. The task consists of filling a list of 136 medical items with the correct values, given a clinical note as input. We reshape the task to make it resemble pure \textbf{classification} and obtain a set of $136$ individual classification tasks, one for each of the original CRF items.
Each classification task is characterized by three or more possible labels. For instance, the classification task defined by the string $s_T$ "\emph{heart rate}" consists in determining whether a patient is "\emph{bradycardic}", "\emph{normocardic}", "\emph{tachycardic}", or if there is not enough information to determine so ("\emph{unknown}"). 
We extended the dataset to 6404 notes\footnote{Can be obtained via email to the corresponding author.}, each annotated for one or more relevant tasks, resulting in $61k$ pairs of note-annotation in total.
Each task has an average of $440$ annotated examples (variance of $147$).
Table~\ref{tab:options} (Appendix) reports all tasks and their label space. 

Furthermore, we define two \textbf{held-out classification tasks} on notes from the SGB dataset, determining if a patient presents any "\emph{chronic}" condition, or determining whether a patient is entering the emergency department with "\emph{dyspnea}" or "\emph{loss of consciousness (tloc)}". 
We reserve the two tasks for evaluating whether the knowledge acquired during \methodname training can transfer to previously unseen tasks.
We believe this setting to be particularly relevant in real-world medical environments, where new classification needs may emerge over time. 
All datasets are described in Table~\ref{tab:data}.

\paragraph{Relevancy Calculation.}
We follow the procedure described in Section~\ref{sec:problem_formulation} to generate the masked training data (Figure~\ref{fig:training_data_creation}). For each task $T$ (out of the $136$), we need to define the target $s_T$ to base the relevance generation on. 
In the datasets we use, each $T$ comes with a short text (one to five words) that defines the dimension to classify the note on (e.g. \emph{heart rate}, Table~\ref{tab:options} for full list). 
Therefore, we can use it as $s_T$, making sure the relevance calculation is based on text that is relevant for $T$ itself. 
For each task, we append $s_T$ to each clinical note in the SGB dataset (template is provided in Appendix~\ref{app:prompts}).

The calculation of the relevance via AttnLRP must be performed over one single generated token; however, all $136$ $s_T$ texts are multi-token sequences. While we could average the relevance calculated for each generated token, this would result in repeated runs over the input. Instead, we use the relevance computed on the middle token of $s_T$ as a computationally efficient proxy. See Appendix~\ref{app:tok_pos_rel} for the validation of this choice.
Furthermore,  for each note, $136$ relevance-calculation prompts can be generated, one per task (defined by $s_T$). Each of these prompts can result in as many training sequences as identified relevance peaks (minus one). To avoid the explosion of training sequences built on the same note, we randomly sample $9$ out of the $136$ possible prompts ($15\%$).
Overall, we run relevance calculation on one million prompts. This step takes $30$ GPU hours on one NVIDIA L40S.

\begin{table*}[ht]
  \centering
  \begin{tabular}{p{2.1cm}|llc|llc|c}
    \toprule
    \multirow{2}{*}{\textbf{Model}}
    &
    \multicolumn{3}{c}{\textbf{SFT}}
    &
    \multicolumn{3}{c}{\textbf{Probing}}
    &
    \multirow{2}{*}{\textbf{\bm{$\Delta_{\mathrm{best}}$}}}
    \\
    \cmidrule(lr){2-4}
    \cmidrule(lr){5-7}
    &
    \textbf{Base}
    &
    \textbf{DSM}
    &
    \textbf{\bm{$\Delta_{\mathrm{DSM}}$}}
    &
    \textbf{Base}
    &
    \textbf{DSM}
    &
    \textbf{\bm{$\Delta_{\mathrm{DSM}}$}}
    &
    \\
    \midrule
     \texttt{Llama3-1B} & 21.9 \scriptsize $\pm$ 1.0& 19.69\scriptsize $\pm$ 1.0    & \textcolor{BrickRed}{-2.2}&     34.1\scriptsize $\pm$  1.1&   47.0 \scriptsize $\pm$ 1.3       & \textcolor{ForestGreen}{+12.9} & \textcolor{ForestGreen}{+25.2}\\ 
     \texttt{Qwen3-1.7B} &  30.9 \scriptsize $\pm$ 1.2 & 22.3 \scriptsize $\pm$ 1.0 & \textcolor{BrickRed}{-8.6}&     33.3 \scriptsize $\pm$ 1.2  & 46.8 \scriptsize $\pm$ 1.3 & \textcolor{ForestGreen}{+13.5} & \textcolor{ForestGreen}{+15.9}\\ 
     \texttt{Gemma3-4B} &37.5 \scriptsize $\pm$ 1.2 & 44.4\scriptsize $\pm$ 1.3     & \textcolor{ForestGreen}{+7.0}&  36.1 \scriptsize $\pm$ 1.2 &  47.5 \scriptsize $\pm$ 1.3      & \textcolor{ForestGreen}{+11.4} & \textcolor{ForestGreen}{+10.0}\\ 
     \texttt{Llama3-8B} & 35.3 \scriptsize $\pm$ 1.0 &42.9 \scriptsize $\pm$ 1.2    & \textcolor{ForestGreen}{+7.6}&  55.6 \scriptsize $\pm$ 1.3 &  55.8  \scriptsize $\pm$ 1.3     & \textcolor{ForestGreen}{+0.2} & \textcolor{ForestGreen}{+20.5}\\ 
     \texttt{Qwen3-8B} &32.5 \scriptsize $\pm$ 1.2 & 40.4 \scriptsize $\pm$ 1.2     & \textcolor{ForestGreen}{+7.9}&  52.5 \scriptsize $\pm$ 1.3 &  60.3 \scriptsize $\pm$ 1.2      & \textcolor{ForestGreen}{+7.7} & \textcolor{ForestGreen}{+27.8}\\
     \bottomrule
  \end{tabular}
  \caption{
    \textbf{Impact of \methodname{} training across task-adaptation strategies.}
    The \textit{SFT} columns report performance for supervised fine-tuning of the base or DSM models, while the \textit{Probing} columns report performance when using a classification head on the final layer.
    For each strategy, $\Delta_{\mathrm{DSM}}$ is the difference between the DSM model and its corresponding base model.
    The final column reports the performance gain of the best DSM adaptation strategy (either SFT or probing) over the standard SFT baseline applied to the base model.
    Results are reported as Macro F1 with $95\%$ confidence intervals.
    }
  \label{tab:all_res}
\end{table*}

\paragraph{Training Sequences Generation.}
Once the relevance is obtained for all notes, we perform Gaussian smoothing (kernel size=$3$, sigma=$1$), and set an upper ($0.4$) and lower ($0.2$) thresholds  to identify the groups of relevant tokens. 
These hyperparameter are selected via manual validation on an held-out set of 100 notes. We set $4$ pairs of lower-upper limit ($0.2/0.4$, $0.2/0.6$, $0.4/0.6$, $0.5/0.6$), and two values for the kernel size of the gaussian smoothing ($1$, $2$). We generate the peaks for each hyperparameter configuration, and count the number of notes where semantically reasonable peaks have been generated (i.e., peaks that are relevant for the target $s_T$), and we select the hyperparameter set that provides the best performance.
Then, we drop all sequences that do not contain at least two peaks, and perform de-duplication, as different $s_T$ texts might have identified the same token groups as relevant for different tasks $T$, which would result in identical training sequences.

For each note containing two relevant spans, we generate training examples by masking one span at a time (Figure~\ref{fig:training_data_creation}, top right). Due to the large number of generated examples, we further filter the dataset by retaining only sequences in which another relevant span occurs to the left of the masked region. This has two reasons: (i) two relevant sequences from the same clinical note most likely contain similar information, and (ii) encouraging the model to infer the masked content from preceding contextual information, thus mimicking the autoregressive nature of the downstream task. 

We collect all the sequences found to be useful for all $136$ tasks in one single dataset of $1.03M$ examples. Each example is composed by a clinical note where a portion has been masked out (to be used as \textit{input}), 
the text $s_T$ used for the masking process, and the text in the masked portion (to be used as \textit{label}). 

\paragraph{\methodname Training.}
To verify if the generated training pairs of input and label can be utilized as sources of knowledge about the downstream classification tasks they were designed for, we analyze three model families in their instructed versions: \texttt{Llama3}, \texttt{Qwen3} in their \scriptsize$\sim$\normalsize\texttt{1B} and \texttt{8B} versions, and \texttt{Gemma3-4B} to represent a mid size between the previous ones.
We train all parameters for one epoch using the hyperparameters described in Appendix~\ref{app:hyperparam} on two NVIDIA H200 GPUs (requiring a total of 68 GPU hours). The learning objective is next token prediction over the prompt structure presented in Appendix~\ref{app:prompts}. 
By doing so, we obtain $5$ \methodname-models, which can be used to analyze the effect of our method.

\section{Evaluation and Discussion}

\paragraph{Downstream Tasks Training.}
We compare the base models with their \methodname versions, to determine the impact of \methodname training. We adapt both versions to the downstream tasks via supervised fine-tuning (SFT) on the $136$ downstream tasks, and determine which performs better .  We divide the CRF dataset into an 85-15 train-test split on an note-item level.
For both settings, we train one single multi-task model for all tasks at once, using LoRA~\citep{hu2022lora}, for one epoch, with the hyperparameters described in Appendix~\ref{app:hyperparam}, for a total of $30$ GPU hours. 

We calculate performance on all $136$ tasks. As aggregation metric to summarize the results on all tasks in one, we use macro F1 following previous work~\citep{ferrazzi2026crf}. We compare the performance of base models which undergo task-oriented SFT to our \methodname-models that undergo the same fine-tuning.
As shown in Table~\ref{tab:all_res} (SFT columns), we observe that \methodname is detrimental for small models ($\sim$\normalsize$1$B), with an average decrease in performance of $-5.4$. On the other hand, it resulted in a homogeneous increase for bigger models, with a $+7.5$ positive impact. 

We hypothesize that this behavior is not due to small models failing to acquire useful knowledge with \methodname, but rather to their limited capacity to coherently generate tokens that reflect this knowledge when prompted to generate the output. In other words, smaller models may successfully encode information that is beneficial for the downstream task, yet lack the representational capacity required to effectively transfer or exploit this knowledge after autoregressive SFT. To verify this hypothesis, further analysis on 
\texttt{\methodname-Llama3-1B} and \texttt{\methodname-Qwen3-1.7B} is required.


\begin{table*}[ht]
  \centering
  \resizebox{\textwidth}{!}{%
  \begin{tabular}{l|ccc|ccc|cc}
    \toprule
    \multirow{2}{*}{\textbf{Model}}
    &
    \multicolumn{3}{c}{\textbf{Tloc-Dyspnea}}
    &
    \multicolumn{3}{c}{\textbf{Chronicity}}
    &
    \multicolumn{2}{c}{\textbf{Average Gain}}
    \\
    \cmidrule(lr){2-4}
    \cmidrule(lr){5-7}
    \cmidrule(lr){8-9}
    &
    \textbf{SFT Base}
    &
    \textbf{CH Base}
    &
    \textbf{CH DSM}
    &
    \textbf{SFT Base}
    &
    \textbf{CH Base}
    &
    \textbf{CH DSM}
    &
    \textbf{\bm{$\Delta_{\mathrm{CH}}$}}
    &
    \textbf{\bm{$\Delta_{\mathrm{SFT}}$}}
    \\
    \midrule
    \texttt{Llama3-1B}  & 50.5 & 79.5 & 84.0 & 65.8 & 65.8 & 69.5 & \textcolor{ForestGreen}{+4.1} & \textcolor{ForestGreen}{+18.6} \\
    \texttt{Qwen3-1.7B} & 36.9 & 78.1 & 86.7 & 68.9 & 65.3 & 68.3 & \textcolor{ForestGreen}{+5.8} & \textcolor{ForestGreen}{+24.6} \\
    \texttt{Gemma3-4B}  & 50.9 & 85.5 & 82.6 & 63.5 & 66.3 & 70.1 & \textcolor{ForestGreen}{+0.4} & \textcolor{ForestGreen}{+19.2} \\
    \texttt{Llama3-8B}  & 71.6 & 80.2 & 81.3 & 64.9 & 64.5 & 68.1 & \textcolor{ForestGreen}{+2.3} & \textcolor{ForestGreen}{+6.5}  \\
    \texttt{Qwen3-8B}   & 47.2 & 79.3 & 80.3 & 56.9 & 64.7 & 68.8 & \textcolor{ForestGreen}{+2.6} & \textcolor{ForestGreen}{+22.5} \\
    \bottomrule
  \end{tabular}}
  \caption{
  \textbf{Transfer to Held Out Classification Tasks.}
  We evaluate whether representations learned through \methodname{} training transfer to classification tasks not used when constructing the \methodname{} training sequences.
  For each task, we report F1 scores for supervised fine-tuning of the base model (\textit{SFT Base}), a classification head trained on the base model (\textit{CH Base}), and a classification head trained on the corresponding \methodname{} model (\textit{CH DSM}).
  $\Delta_{\mathrm{CH}}$ is the average F1 gain of CH DSM over CH Base across the two tasks.
  $\Delta_{\mathrm{SFT}}$ is the average F1 gain of CH DSM over SFT Base across the two tasks.
  }
  \label{tab:ood_tasks}
\end{table*}
\paragraph{Probing.}
To better understand why training with \methodname improves downstream performance only for bigger models, we perform a probing analysis by training a shallow classifier on top of the last hidden layer of the small models ($\sim$ 1B). 
We add a classification head on the last hidden representation, composed by a linear layer (output size of $256$), a ReLU activation with $0.1$ dropout, and an output layer to map the $256$ neurons to the output classes. We train  this head with the hyperparameters described in Appendix~\ref{app:hyperparam}, for an overall of $24$ L40S GPU hours. 
The results are reported in Table~\ref{tab:all_res} (last four columns). 
We observe a consistent positive effect, suggesting that \methodname successfully injects useful information into the learned representations, even in cases where this does not translate into improved downstream autoregressive generation performance. The models that underwent our training result in a homogeneous average increase of $+13.2$ points.

\paragraph{From SFT to Probing.}
The probing analysis suggests that even smaller models learn an useful representation for the downstream tasks. Furthermore, by comparing the results obtained via SFT and probing, we observed that the potential of the models on these tasks goes much beyond what they exhibit after standard, auto-regressive SFT (an effect that has been analyzed by~\citet{LYU2026114341}). In fact, \texttt{Llama3-1B} with a classification head before SFT achieves a Macro F1 $+11.2$ points higher than after standard SFT and without the head  ($+2.4$ for \texttt{Qwen3-1.7B}). 
Building on this evidence, we train a classification head on top of the larger models as well, to verify if they could achieve better performance. We use the same structure and training parameters described above, obtaining the results shown in Table~\ref{tab:all_res}. 

We highlight two aspects: first, when adapting models to downstream tasks by adding a classification head, \methodname gives an average improvement of $+9.1$ in Macro F1, showcasing that the training procedure effectively injects useful knowledge into the models' representations from the unannotated data. Second, the average performance gain moving from auto-regressive SFT on the base model to classification heads on the \methodname version is $+19.9$ points, showing this approach to be the most effective.

\paragraph{Held Out Tasks.}
We reserve two tasks to evaluate whether the knowledge acquired during \methodname training on the 136 medical classification tasks can transfer to new classification settings.
We employ two datasets \textit{Chronicity} and \textit{T-D} described in Section~\ref{sec:setup} and in Table~\ref{tab:data}. The evaluation metric for both tasks is the F1 score.

We build on the finding that classification heads overperform SFT, and follow this to adapt base models and their \methodname counterparts. In addition, we provide for comparison standard SFT results on the base models, and report all findings in Table~\ref{tab:ood_tasks}. The probing strategy showcases that \methodname results in higher F1 ($+3.0$ on average across all models) when compared to the base, and it presents an even higher increase of $+18.3$ when compared with the base with SFT.

\paragraph{Baselines.}
We compare \methodname to other methods, to contextualize its efficacy.
The natural baseline for text classification tasks are \textbf{encoder-only models}. We fine-tune on the $136$ CRF downstream tasks \texttt{BERT} and its multilingual version \texttt{multilingual-BERT}~\citep{devlin-etal-2019-bert}, \texttt{BioClinicalBERT}~\citep{alsentzer-etal-2019-publicly}, \texttt{ModernBERT}~\citep{warner-etal-2025-smarter}, and a version of BERT explicitly trained for Italian \citep{stefan_schweter_2020_4263142}. 
The models are trained to receive as input a clinical note and the text describing the task $s_T$ (e.g., "\emph{heart rate}") with the hyperparameters defined in Appendix~\ref{app:hyperparam}. 

Moreover, we provide the performance of the decoder model that showed the best results in our experiments (\texttt{Qwen3-8B}), prompted \textbf{0- and 2-shot}. 
For 0-shot, we run inference $3$ times with slightly different system prompts  in order to account for instabilities, following~\citet{ICLR2024_6c0e99d7}, and report the average.
For 2-shot, we randomly sample the examples, and run inference $3$ times to mitigate bias due to their choice.
Results are presented in Table~\ref{tab:baselines}, showing preference for encoders (up to $+13.4$ points), which are still outperformed by \methodname.

\paragraph{State of the Art Comparison.}
Our approach relies on training on the unannotated, raw data. As described in Section~\ref{sec:related_work}, the typical approach to expose models to large amounts of raw data is \textbf{continual pretraining} (CPT), whose underlying hypothesis inspires \methodname. We then perform CPT on a model on the same (unmasked) raw clinical notes included in our method. We train using the same hyperparameters as in \methodname, with the standard next-token-prediction objective.  While CPT has a positive impact, it still falls short w.r.t. \methodname ($-6.3$ points).

Finally, we include a standard \textbf{self-learning} approach, which we take inspiration from to design the training objective of Equation~\ref{eq:self-sup}. We train a model on the downstream tasks via SFT on the annotated CRF data (similar to the SFT Base models from Tables~\ref{tab:all_res} and~\ref{tab:ood_tasks}), and then use it to generate synthetic labels for the unannotated SGB dataset. Then, we train the original base model on the combination of the original and the synthetic labeled data via SFT. While outperforming the base model adapted to the task via SFT ($+15.3$ points), it still falls behind our approach ($-12.5$).

Overall, these results highlight that our approach obtains better performance than previous methods, by effectively integrating knowledge embedded in raw data and shaping it through a procedure that enhances classification performance.
Moreover, they disentangle the sources of \methodname's improvements. While exposure to in-domain clinical data is itself beneficial (as shown by CPT), \methodname\ reveals an additional gain attributable to the task-oriented representations learnt by our method. Together with the benefits of exposure to task-related examples, these results suggest that the observed improvements arise not simply from additional data, but from how unlabeled data are transformed into task-relevant training signals.

\paragraph{Two peaks and relevance attribution ablation.}
The method we present relies on the "two peaks" criterion (Section~\ref{sec:problem_formulation}), meaning that \methodname models are trained on sequences that include at least two distinct groups of tokens which contain information relevant to the downstream task. To quantify the impact of this theoretical choice, we ablate on the number of peaks, training the best performing model (\texttt{Qwen3-8B}) on sequences exhibiting one single peak (i.e., one single group of relevant tokens). 
It must be noted that notes with only one peak are less than the ones with more. To account for it, we train on five epochs, reaching comparable training size.
We observed that limiting the training to such sequences leads to worse results, with a drop in Macro F1 of $-13.1$ (\texttt{Qwen3-8B + one peak + CH} in Table~\ref{tab:baselines}), experimentally motivating the "two peaks" choice.

In addition, we provide results obtained by ablating the relevance attribution step, where we train the model to autoregressively reconstruct the sequences that respected the two peaks criterion, but using randomly masked portion of text instead of the ones found via AttnLRP. Table~\ref{tab:baselines} (\texttt{Qwen3-8B + random mask}) shows that the removal of this component has a negative impact ($-3.9$ in Macro F1). This ablation highlights that the effectiveness of the signal embedded in the auto-regressive sequence reconstruction depends on the quality of the relevance attribution method.

\begin{table}[t]
  \centering
  \resizebox{\columnwidth}{!}{%
  \begin{tabular}{l|c}
    \toprule
     \textbf{Model} &  \multicolumn{1}{c}{\textbf{Macro F1}} \\ \toprule
     \texttt{ClinicalBERT} & 34.1 \scriptsize $\pm$ 1.2  \\
     \texttt{multilingual-BERT} &   34.2 \scriptsize $\pm$ 1.2  \\ 
     \texttt{BERT} &         36.8 \scriptsize $\pm$ 1.2    \\
     \texttt{BERT-Italian} &   38.2 \scriptsize $\pm$ 1.3  \\ 
     \texttt{ModernBERT} &   44.1 \scriptsize $\pm$ 1.3  \\  \midrule
     \texttt{Qwen3-8B  + 0-shot} &  27.7 \scriptsize $\pm$ 0.6 \\
     \texttt{Qwen3-8B   + 2-shot} & 30.7 \scriptsize $\pm$ 0.7 \\ 
     \texttt{Qwen3-8B   + SFT} & 32.5 \scriptsize $\pm$ 1.1 \\
     \texttt{Qwen3-8B   + CH} & 52.5 \scriptsize $\pm$ 1.1 \\  \midrule
     \texttt{Qwen3-8B   + CPT + SFT}  & 33.5 \scriptsize $\pm$ 1.3 \\ 
     \texttt{Qwen3-8B   + CPT + CH}  & 54.0 \scriptsize $\pm$ 1.4 \\ \midrule
     \texttt{Qwen3-8B   + synthetic SFT } & 47.8 \scriptsize $\pm$ 1.3 \\ \midrule
     \texttt{Qwen3-8B  + random mask + SFT } & 30.4 \scriptsize $\pm$ 1.2 \\
     \texttt{Qwen3-8B  + random mask + CH}  & 56.4 \scriptsize $\pm$ 1.1 \\ \midrule
     \texttt{Qwen3-8B  + one peak + SFT}  &  30.0 \scriptsize $\pm$ 1.2 \\
     \texttt{Qwen3-8B  + one peak + CH}  &  47.2 \scriptsize $\pm$ 1.3 \\  \midrule
     \methodname-\texttt{Qwen3-8B-CH} & \textbf{60.3} \scriptsize $\pm$ 1.2 \\
   \bottomrule
  \end{tabular}}
  \caption{\textbf{Comparison to baselines and state-ot-the-art} for the $136$ CRF tasks compared with our best performing model (\methodname-\texttt{Qwen3-8B-CH} with a classification head \texttt{CH}). We provide results for $0$- and $2$-shot; continual pretraining (\textit{CPT}); self-learning where the model is trained via SFT on synthetic labels generated by itself (\textit{synthetic SFT}). We also showcase the performance of applying \methodname with \textit{random mask}ing instead of relying on relevance attribution, and of ablating on the "two peaks" criterion (\textit{one peak}).}
  \label{tab:baselines}
\end{table}

\begin{figure*}
    \centering
    \includegraphics[width=\textwidth]{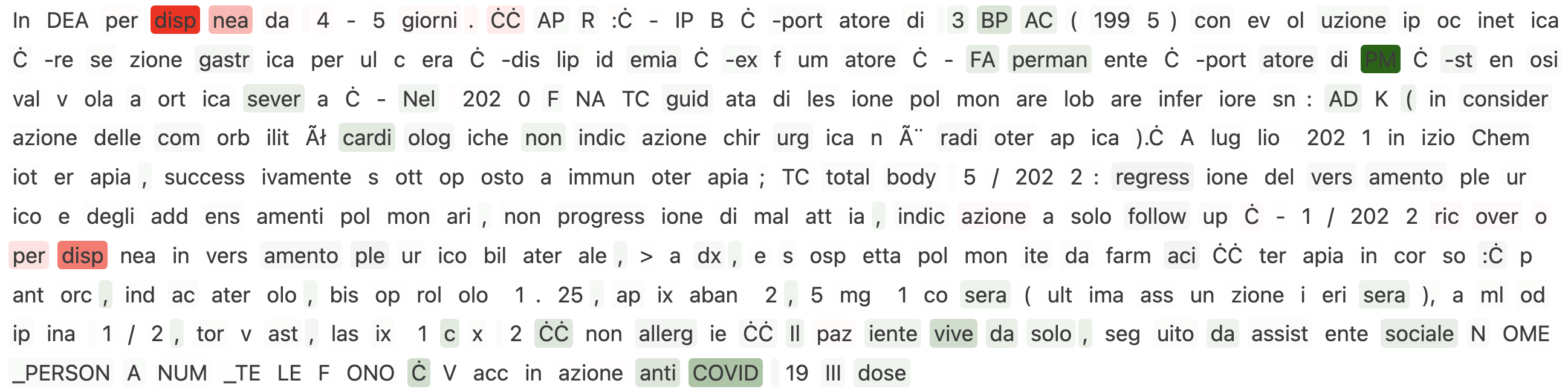}
    \caption{\textbf{Impact of the \methodname training on the relevance attributed by AttnLRP to the input for the task \textit{"Presence of pacemaker Pacemaker"}.} Red represents higher relevance being assign to a specific token by the base model when compared to its \methodname version, while green the opposite. While the base model primarily focuses on terms related to \textit{dyspnea}, the \methodname model shifts its attention toward mentions of "\textit{PM}" (pacemaker), which is substantially more relevant to the target condition}
    \label{fig:ex_pacemaker}
\end{figure*}

\paragraph{\methodname further analysis.}
Our training procedure modifies the model's parameters to encode information extracted from unlabeled data, effectively steering the model's internal representations toward task-relevant patterns present in the raw corpus. To better understand the effect of this training procedure, we analyze how it alters the relevance assigned to different portions of the input. 
More specifically, we provide the model with the same prompt used during the relevance extraction phase of our pipeline,
and compute the relevance attributed to the input tokens while generating the target item using the same AttnLRP score used when constructing the dataset.

This analysis reveals notable qualitative differences between the base model and the corresponding \methodname variant. In several cases, our models assign higher relevance to portions of text that are semantically more aligned with the target concept (e.g, Figures~\ref{fig:ex_pacemaker}
and ~\ref{fig:ex_app1}
-\ref{fig:ex_app4} in Appendix).

We quantify this effect by leveraging the labeled clinical notes, where annotators also identified the portions of text considered relevant for each of the 136 tasks. We measure the change in relevance assigned by AttnLRP to these ground-truth spans when moving from the base model to the corresponding \methodname variant. For \texttt{Qwen-8B}, we observe an average relevance increase of $37.2\%$ over the annotated spans, suggesting that the proposed self-supervised training objective effectively steers the model towards assigning higher importance to actually relevant portions of the input.

\section{Conclusion}
\label{sec:conclusion}

In this work, we introduce \methodname, a task-oriented self-supervised framework to leverage unlabeled, domain-specific corpora to improve the downstream classification performance of decoder-only models, combining the assumptions underlying \textit{self-learning} and \textit{continual pretraining}. 

First, we propose to use input attribution methods to mask out task-relevant portions of unlabeled text, and train to autoregressively reconstruct them, encouraging models to learn domain-specific patterns from raw data, guided by relations identified as relevant to the target tasks by models themselves.

Second, we evaluate \methodname\ on $136$ tasks derived from 1.9M clinical notes. We adapt models to the tasks with a classification head, as probing experiments show consistent improvement over standard SFT across model families and scales ($+19.9$ F1 Macro points), also showing similar effectiveness on two held-out tasks ad well.

Third, we compare \methodname to standard \textit{continual pretraining} ($+6.3$) and \textit{self-learning} approaches ($+12.5$), and propose an ablation study on two  components of the method to further highlight the soundness and effectiveness of our approach.


Overall, \methodname contributes a novel way of exploiting unlabeled data for decoder-only classification, by combining ideas from masked training, continual pretraining, and self-learning. In contrast to conventional continual pretraining, it performs task-guided adaptation from unlabeled data, while, unlike existing decoder-only masking approaches, it preserves the standard autoregressive objective and the model’s generative capabilities. Rather than relying on synthetic labels, \methodname uses the model’s own relevance estimates to identify task-relevant spans and turns them into self-supervised training examples. In this way, it exploits the decoder’s prior knowledge of the relationship between the domain and the downstream task, providing a targeted mechanism for learning task-relevant dependencies from unlabeled text. The ablation results further support the central role of relevance-guided masking, showing that learning from the dependencies the model itself considers relevant is particularly effective to acquire task-specific knowledge.



\section*{Limitations}

Our experiments are limited to Italian clinical notes and relatively small decoder-only models, although results suggest that larger models may benefit more from \methodname. Due to the computational cost of AttnLRP, we train on a subset of the original 1.9M-note corpus and perform joint self-supervised training across all tasks rather than task-specific training. Additionally, the proposed framework assumes that downstream tasks can be represented through short textual descriptions $s_T$. Future work should explore more efficient relevance extraction methods, alternative masking objectives, and extensions to broader task settings and larger models.

\section*{Ethical considerations}

Although our approach aims to reduce the dependency on expensive manual annotation, models trained on medical data may still inherit biases, incompleteness, or inaccuracies present in the underlying clinical records. Consequently, the proposed models should not be considered a substitute for professional medical judgment.

Additionally, relevance attribution methods may produce imperfect or misleading explanations of model behavior. Therefore, the relevance analyses presented in this work should be interpreted as approximations of model reasoning rather than faithful causal explanations.

\section*{Acknowledgments}
This work has been partially funded by the European Union under the Horizon Europe eCREAM Project (Grant Agreement No.101057726). Views and opinions expressed are however those of the authors only and do not necessarily reflect those of the European Union or the European Health and Digital Executive Agency (HADEA). Neither the European Union nor the granting authority can be held responsible for them.


\bibliography{custom}

\appendix

\section{Relevance Computation Details}
\label{app:tok_pos_rel}

\paragraph{Encoder-only Masking Strategies.}
The method we present is inspired by language masking training strategies used for encoder-only models. It has been shown that the choice of what portions of text to mask is important when training encoders using masked language modeling~\citep{gu-etal-2020-train,ye-etal-2021-influence,sadeq-etal-2022-informask, golchin-etal-2023-mask}. We take inspiration from this paradigm, shifting it to autoregressive generative models.

\paragraph{Middle Token and AttnLRP Choices.}

As the cost of calculating the relevance using AttnLRP for all tokens in the output increases linearly with the length of the output, we approximate the average relevance if the input on a certain output using one single output token.
A practical challenge arises when the target label is composed of multiple tokens: relevance can be computed using the hidden states associated with different decoding steps, potentially leading to different attribution patterns. In particular, it is unclear whether relevance should be estimated using the first generated token, the last one, or by aggregating information across all label tokens.

To investigate this aspect, we conduct a validation experiment on a medical Named Entity Recognition (NER) dataset \cite{magnini2020e3c}. For each annotated entity, we construct a prompt of the form:
\begin{promptbox}
   You are an entity extractor. \\
\{text\} \{label\}
\end{promptbox}
and evaluate whether the relevance assigned by the model to the tokens in \texttt{\{text\}} corresponding to the target entity is significantly higher than what would be expected under a uniform relevance distribution.

\begin{figure}[t]
  \includegraphics[width=0.48\linewidth]{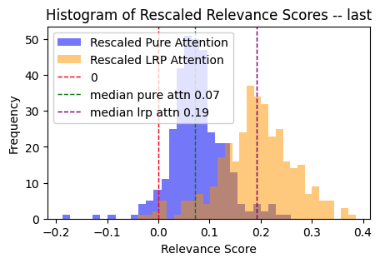} \hfill
  \includegraphics[width=0.48\linewidth]{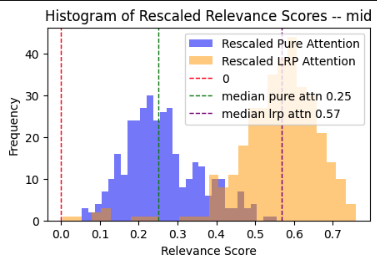} \hfill
  \includegraphics[width=0.48\linewidth]{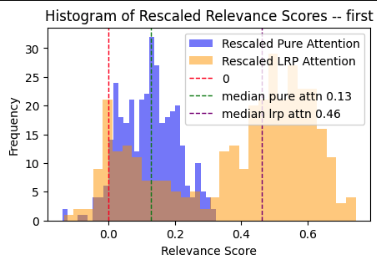}\hfill
  \caption {\textbf{Comparison between the relevance attributed to ground truth portions of text} i) by \textit{Pure Attention} (i.e., average attention across all layers), and AttnLRP, ii) when utilizing different tokens of $s_T$ for the relevance calculation. The $x$ axis represents the increase in relevance that each method is assigning to the ground truth textual portion with respect with uniform distribution (i.e, the more shifted to the right, the better). The best performing configuration is using the \textit{middle} token with AttnLRP (top-right histogram).}
\label{fig:validation_ner}
\end{figure}

In parallel, we compare two relevance estimation methods, the average of the attention scores among all layers, and AttnLRP, by computing token relevance using the hidden states associated with: (i) the first token of the label, (ii) the last token of the label, and (iii) the middle token of the label. Our results show that using the middle token consistently produces the strongest localization signal, yielding higher relevance scores over the entity span (Figure~\ref{fig:validation_ner}). This suggests that intermediate decoding steps provide a more stable representation of the semantic content of multi-token labels compared to boundary tokens.
By doing so, we also validate the usage of AttnLRP, which shows the best performance in terms of relevance attribution to the ground truth in the original text.

\section{Training parameters}
\label{app:hyperparam}

\paragraph{\methodname training}
We train using \texttt{deepspeed}  \texttt{Zero2} and \texttt{flash} \texttt{attention}, on an effective batch size of $128$, using AdamW optimizer with a learning rate of $4\cdot10^{-5}$ (cosine scheduler), weight decay of $0.001$, warmup ratio $0.1$.

\paragraph {Supervised fine-tuning}
For Supervised fine-tuning, we use a batch size of $64$, learning rate of $3 \cdot 10^{-4}$, warmup ratio $0.1$, using AdamW optimizer with cosine scheduler and epsilon $1 \cdot 10^{-7}$. We use LoRA~\citep{hu2022lora}, with $r$ $32$, alpha $64$, and dropout $0.05$. We construct SFT sequences according to the  template in Appendix~\ref{app:prompts}.

\paragraph{Probing}
We train for $20$ epochs, with learning rate $1 \cdot 10^-3$ and the AdamW optimizer.

\paragraph{Encoder-only baseline}
Training of encoder-only baselines is performed for $10$ epochs, with an AdamW optimizer, and a learning rate of $2\cdot10^{-5}$. Initial experiments showed that increasing the number of epochs would not provide any benefit.

\section{Prompts}
\label{app:prompts}

\paragraph{Relevance calculation}
We compute the relevance by using the following prompt template:
\begin{promptbox}
\{clinical note\}\\
Given the patient’s medical history, the following is observed \{$s_T$\}.
\end{promptbox}

Note that models' special tokens are added where required --as well as the added text -- but they are excluded from the relevance calculation, following the findings of~\citet{ICLR2024_5e5fd18f}.

\paragraph{\methodname training}
The self-training is performed using the following template:
\begin{promptbox}
Fill in the masked word in the following sentence.\\
\{clinical note\} 
\{label\}
\end{promptbox}

Note that the simple system prompt is provided to leverage models instruction following, and models' special tokens are added where required to determine the beginning and end of the instruction and output.

\paragraph{Supervised fine-tuning}
When fine-tuning \methodname models on the downstream CRF classification tasks (each task identified by a short description $s_T$), we combine the data examples according to the following template:
\begin{promptbox}
Fill in the masked word in the following sentence. \\
\{masked clinical note\}\\
<crf\_item> \{{$s_T$}\}? \{masking token\} <\textbackslash crf\_item> <options>\{options\}<\textbackslash options>\\

\{label\}
\end{promptbox}
The masking token is a model-dependent special token that is used to substitute the masked text.

On the other hand, when fine-tuning base models (e.g., \texttt{Qwen/Qwen3-8B}) on the same tasks, we initially kept a different template, removing the masking token. Then, we observed that changing the prompt did not impact the performance, and kept the previous versions for all. This finding is consistent with~\citet{LYU2026114341}.

\begin{table*}
\centering
\scriptsize
\begin{tabular}{p{7cm}p{6cm}}
\toprule
\textbf{options} & \textbf{items} \\
\midrule
A; V; P; unknown & ['level of consciousness'] \\ \hline
bradycardic; normocardic; tachycardic; unknown & ['heart rate'] \\ \hline
bradypneic; eupneic; tachypneic; unknown & ['respiratory rate'] \\ \hline
certainly active; possibly active; certainly not active; unknown & ['active neoplasia'] \\ \hline
certainly chronic; possibly chronic; certainly not chronic; unknown & ['chronic pulmonary disease', 'chronic respiratory failure', 'chronic cardiac failure', 'chronic renal failure', 'chronic metabolic failure', 'chronic rheumatologic disease', 'chronic dialysis'] \\ \hline
current; past; unknown & ['presence of respiratory distress'] \\ \hline
hypotensive; normotensive; hypertensive; unknown & ['blood pressure'] \\ \hline
hypothermic; normothermic; hyperthermic; unknown & ['body temperature'] \\ \hline
measured; unknown & ['spo2', 'ph', 'pa02', 'pac02', 'hc03-', 'lactates', 'hemoglobin', 'platelets', 'leukocytes', 'c-reactive protein', 'blood sodium', 'blood potassium', 'blood glucose', 'creatinine', 'transaminases', 'inr', 'troponin', 'bnp or nt-pro-bnp', 'd-dimer', 'blood calcium', 'serum creatinine kinase', 'blood alcohol', 'blood drug dosage', 'urine drug test'] \\ \hline
pos; neg; unknown & ['carotid sinus massage', 'supine-to-standing systolic blood pressure test', 'blood in the stool', 'sars-cov-2 swab test'] \\ \hline
short; long; unknown & ["duration of the patient's consciousness recovery", "duration of the patient's unconsciousness"] \\ \hline
walking independently; walking with auxiliary aids; walking with physical assistance; bedridden; unknown & ['level of autonomy (mobility)'] \\ \hline
y; n; unknown & ['first episod of epilepsy', 'known history of epilepsy', 'history of allergy', 'history of recent trauma', 'pregnancy', 'history of drug abuse', 'history of alcohol abuse', 'anticoagulants or antiplatelet drug therapy', 'presence of prodromal symptoms', 'compliance with antiepileptic therapy', 'tloc during effort', 'tloc while supine', 'antiepileptic therapy already in place', 'drowsiness, confusion, disorientation as postcritical state', 'stiffness during the episode', 'drooling during the episode', 'tonic-clonic seizures', 'poly-pharmacological therapy', 'pale skin during the episode', 'eye deviation during the episode', 'diffuse vascular disease', 'neuropsychiatric disorders', 'presence of pacemaker', 'presence of defibrillator', 'cardio-pulmonary resuscitation', 'antihypertensive therapy', 'cardiovascular diseases', 'neurodegenerative diseases', 'peripheral neuropathy', 'immunosuppression', 'palliative care', 'situation description, like coughing, prolonged periods of straining, sudden abdominal pain, phlebotomy', 'problematic family context', 'need but absence of a caregiver', 'homelessness', 'living alone', 'chest pain', 'head or other districts trauma', 'tongue bite', 'agitation', 'foreign body in the airways', 'improvement of dyspnea', 'presence of dyspnea', 'dementia', 'general condition deterioration', 'ab ingestis pneumonia', 'further seizures in the ed', 'improvement of patient’s conditions', 'neurologist consultation', 'ecg, any abnormality', 'ecg monitoring, any abnormality', 'eeg, any abnormality', 'thoracic ultrasound, any abnormalities', 'chest rx, any abnormalities', 'gastroscopy , any abnormalities', 'brain ct scan, any abnormality', 'brain mri, any abnormality', 'cardiac ultrasound, any abnormality', 'chest ct scan, any abnormality', 'pulmonary scintigraphy, any abnormality', 'abdomen ct scan, any abnormality', 'compression ultrasound (cus), any abnormality', 'performance of thoracentesis', 'administration of diuretics', 'administration of steroids', 'administration of bronchodilators', 'administration of oxygen/ventilation', 'blood transfusions', 'administration of fluids', 'heart failure', 'pneumonia', 'copd exacerbation', 'acute pulmonary edema', 'asthma exacerbation', 'respiratory failure', 'intoxication', 'covid 19', 'influenza and various infections', 'pneumothorax', 'situational syncope', 'epilepsy / epileptic seizure', 'pulmonary embolism', 'arrhythmia', 'cardiac tamponade', 'aortic dissection', 'acute coronary syndrome', 'hemorrhage', 'severe anemia', 'concussive head trauma'] \\ 
\bottomrule
\end{tabular}
\caption{\textbf{List of the $136$ Case Report Form tasks and their respective labels' space (\textit{options}).} Several tasks share the same labels. The elements in \textit{items} are used to define the classification tasks, and as targets $s_T$ to quantify the relevance during the \methodname relevance attribution step.}
\label{tab:options}
\end{table*}

\begin{figure*}
    \centering
    \includegraphics[width=\textwidth]{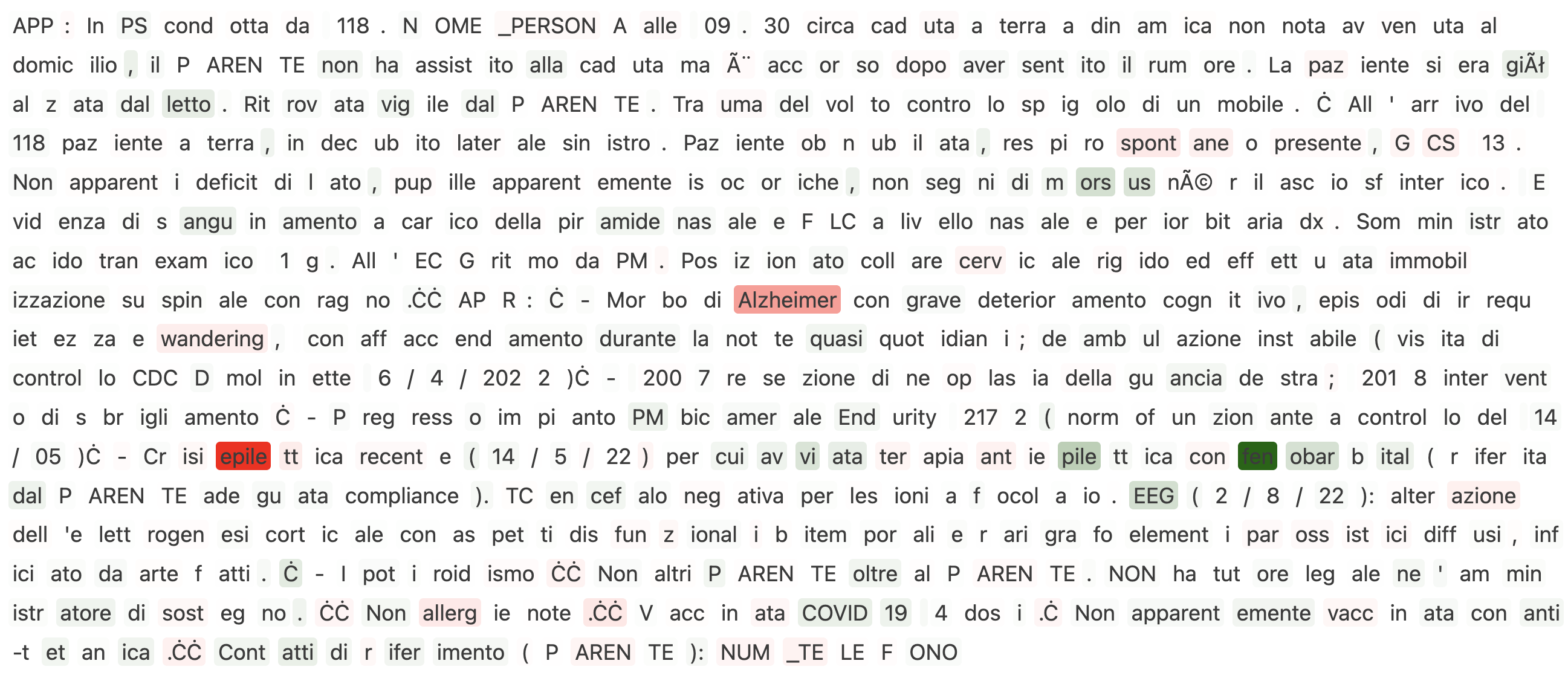}
    \caption{\textbf{Impact of the \methodname training on the relevance attributed by AttnLRP to the input when generating the text \textit{"Antiepileptic therapy already in place"}.} Red represents higher relevance to be assign to a specific token by the base model when compared to its \methodname version, while green the opposite. Our model assigns much more importance to "fenobarbital", whichis an antiepileptic drug, and reduces it to Alzheimer (irrelevant here). Interestingly, relvenace is lowered for "Crisi epilettica" (i.e., epileptic crisis), suggesting that the model has learnt to move attention to plausible portions to the ones containing the actual information. Epileptic crisis is obviously linked to its therapy, but gives no factual knowledge about medications being in place already.}
    \label{fig:ex_app1}
\end{figure*}

\begin{figure*}
    \centering
    \includegraphics[width=\textwidth]{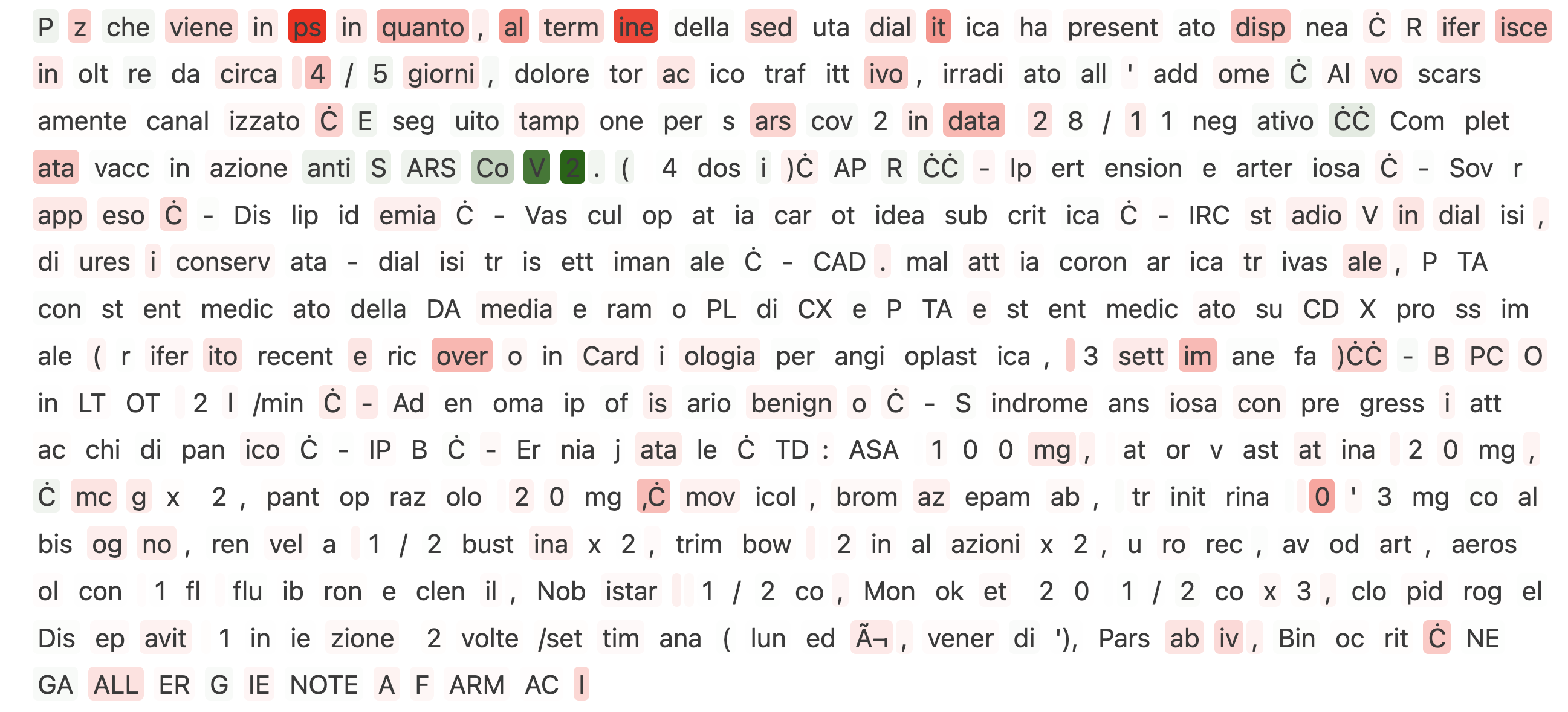}
    \caption{\textbf{Impact of the \methodname training on the relevance attributed by AttnLRP to the input when generating the text \textit{"SARS-CoV-2 swab test"}.} Red represents higher relevance to be assign to a specific token by the base model when compared to its \methodname version, while green the opposite. Our model assigns much more importance to the token "Anti SARS CoV2", and reduces it to other tokens that are indeed not required by the target. }
    \label{fig:ex_app2}
\end{figure*}

\begin{figure*}
    \centering
    \includegraphics[width=\textwidth]{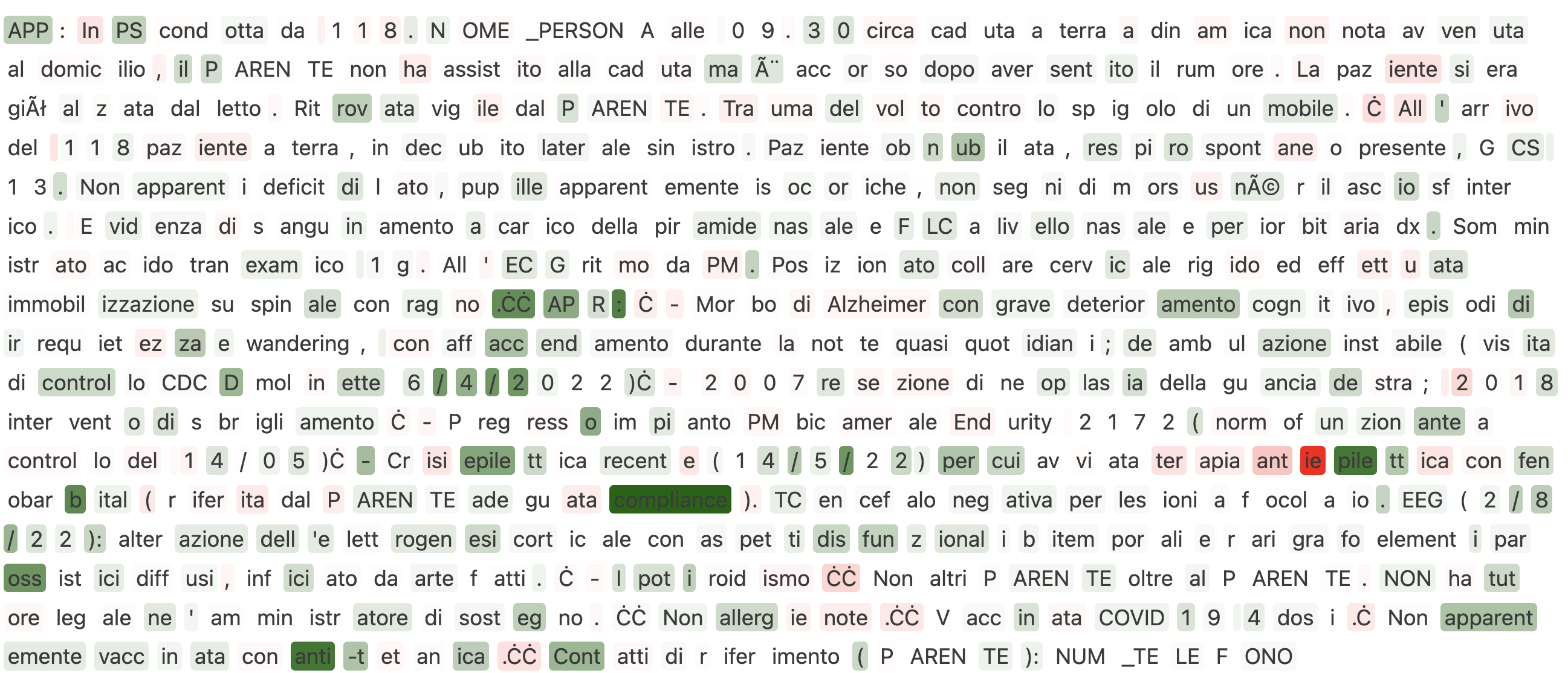}
    \caption{\textbf{Impact of the \methodname training on the relevance attributed by AttnLRP to the input when generating the text \textit{"Compliance with antiepileptic therapy"}.} Red represents higher relevance to be assign to a specific token by the base model when compared to its \methodname version, while green the opposite. Our model assigns much more importance to the tokens "compliance", "fenobarbital", which are both relevant to the target. }
    \label{fig:ex_app3}
\end{figure*}

\begin{figure*}
    \centering
    \includegraphics[width=\textwidth]{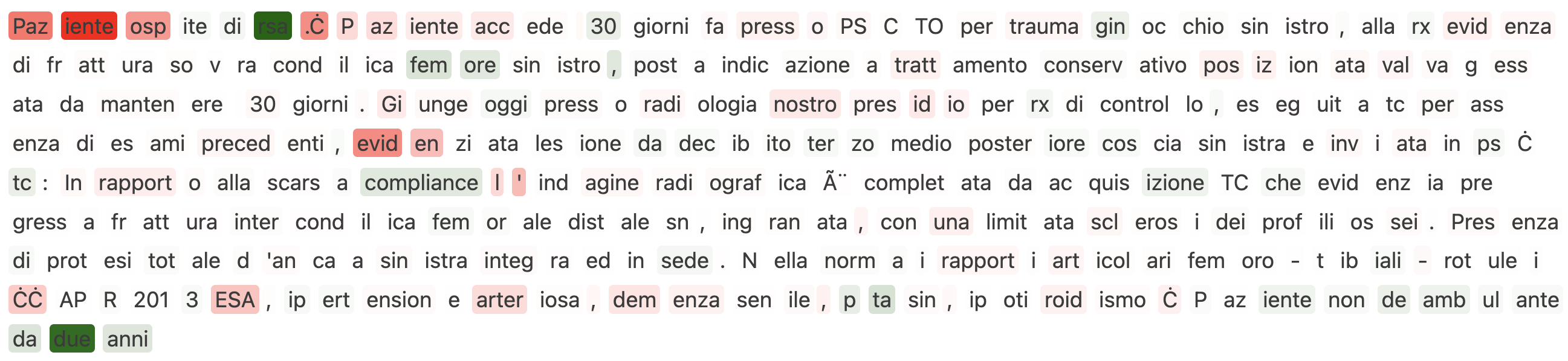}
    \caption{\textbf{Impact of the \methodname training on the relevance attributed by AttnLRP to the input when generating the text \textit{"level of autonomy"}.} Red represents higher relevance to be assign to a specific token by the base model when compared to its \methodname version, while green the opposite. Our model assigns much more importance to the tokens "rsa" (i.e., care home) and "Paziente non deambula da due anni" (The patient has not been able to walk for two years). }
    \label{fig:ex_app4}
\end{figure*}

\end{document}